\documentclass[10pt,twocolumn,letterpaper]{article}

\usepackage{iccv}
\usepackage{times}
\usepackage{epsfig}
\usepackage{graphicx}
\usepackage{amsmath}
\usepackage{amssymb}

\DeclareMathOperator*{\argmin}{arg\,min}

\newcommand{\D}{\mathcal{D}}

\usepackage[ruled, vlined]{algorithm2e}
\usepackage{graphicx}
\usepackage{amsmath}
\usepackage{amssymb}
\usepackage{booktabs}
\usepackage{amsfonts} 

\usepackage{float}
\usepackage{subfig}

\usepackage[pagebackref=true,breaklinks=true,letterpaper=true,colorlinks,bookmarks=false]{hyperref}

\iccvfinalcopy 

\def\iccvPaperID{6663} 

\ificcvfinal\fi

\begin{document}

\title{Visual Prompt Based Personalized Federated Learning}

\author{
Guanghao Li\textsuperscript{\rm 1,*}
\quad 
Wansen Wu\textsuperscript{\rm 1,}\thanks{Co-first authors} 
\quad 
Yan Sun\textsuperscript{\rm 2}
\quad 
Li Shen\textsuperscript{\rm 3,}\thanks{Corresponding author: Li Shen} 
\quad
Baoyuan Wu\textsuperscript{\rm 4}
\quad 
Dacheng Tao\textsuperscript{\rm 2,3}
\\
\textsuperscript{\rm 1}National University of Defense Technology, China; 
\textsuperscript{\rm 2}The University of Sydney, Australia\\
\textsuperscript{\rm 3}JD Explore Academy, China; 
\textsuperscript{\rm 4}The Chinese University of Hong Kong, Shenzhen, China\\
{\tt\small \{lgh, wuwansen14\}@nudt.edu.cn; woodenchild95@outlook.com}\\
{\tt\small \{mathshenli,wubaoyuan1987,dacheng.tao\}@gmail.com
}
}

\maketitle
\ificcvfinal\thispagestyle{empty}\fi

\begin{abstract}
As a popular paradigm of distributed learning, personalized federated learning (PFL) allows personalized models to improve generalization ability and robustness by utilizing knowledge from all distributed clients. Most existing PFL algorithms tackle personalization in a model-centric way, such as personalized layer partition, model regularization, and model interpolation, which all fail to take into account the data characteristics of distributed clients. In this paper, we propose a novel PFL framework for image classification tasks, dubbed pFedPT, that leverages personalized visual prompts to implicitly represent local data distribution information of clients  and provides that information to the aggregation model to help with classification tasks. Specifically, in each round of pFedPT training, each client generates a local personalized prompt related to local data distribution. Then, the local model is trained on the input composed of raw data and a visual prompt to learn the distribution information contained in the prompt. During model testing, the aggregated model obtains prior knowledge of the data distributions based on the prompts, which can be seen as an adaptive fine-tuning of the aggregation model to improve model performances on different clients. Furthermore, the visual prompt can be added as an orthogonal method to implement personalization on the client for existing FL methods to boost their performance. Experiments on the CIFAR10 and CIFAR100 datasets show that pFedPT outperforms several state-of-the-art (SOTA) PFL algorithms by a large margin in various settings.

\end{abstract}

\section{Introduction}
\begin{figure*}
    \centering
    \includegraphics[width=0.75\textwidth]{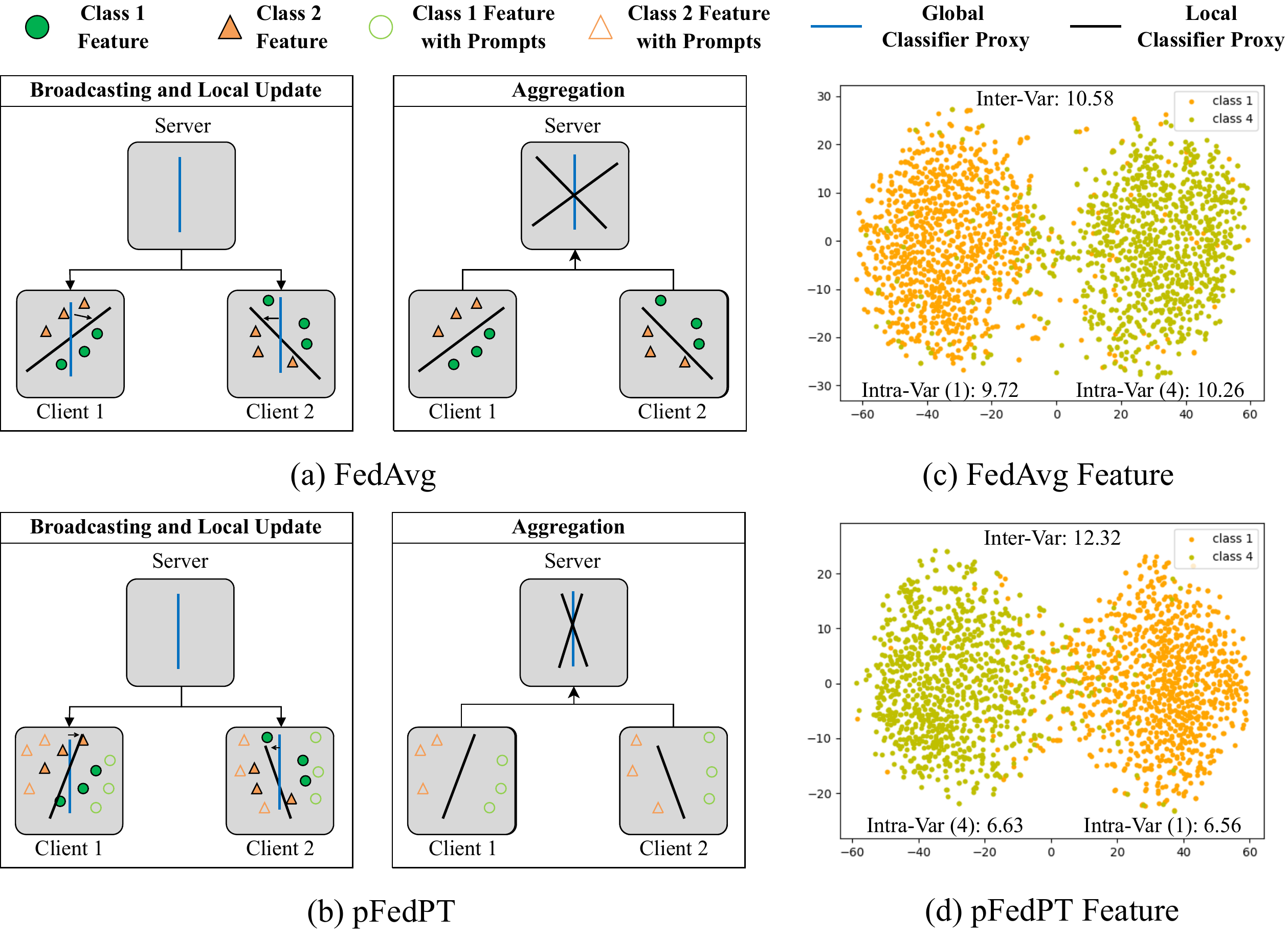}
    \vspace{-0.8em}
    \captionsetup{font=small}
    \caption{Differences in local update and aggregation phases between FedAvg and pFedPT. In the figure, the lines represent the decision boundaries defined by the backbone. Assume that each client has two classes represented by different shapes. (a) In FedAvg, due to the heterogeneity of data in each client, the significant difference in local updates affects the final model aggregation. However, (b) the pFedPT adds personalized visual prompts to the client data, which change the original data characteristics and improve the fit of the backbone on the client. (c) and (d) are the t-SNE visualization results of the final hidden layer trained by FedAvg and pFedPT on the client with only classes 1 and 4. pFedPT increases the inter-class variation (Inter-Var) and decreases the intra-class variation (Intra-Var).}
    \label{fig:dif}
    
\end{figure*}

Personalized federated learning (PFL)~\cite{deng2020adaptive,huang2022achieving,dai2022dispfl} is a novel paradigm proposed to overcome the impacts of heterogeneity across isolated clients. Instead of training a single aggregated model like in  Federated learning(FL)~\cite{mcmahan2017communication,tan2022towards,li2022fedhisyn,gao2022feddc}, PFL generates a personalized local model on each client that is more in line with the local data distribution by jointly considering the aggregated model and the personalized data. There are two main challenges lying in PFL. One is how to extract useful global features from models trained on each local heterogeneous dataset. The other is how to incorporate the extracted global features with the personalized features, yielding a better client-specific model.


Several works have been proposed to address the above challenges from a model perspective. PFL algorithms with a decoupling model ~\cite{arivazhagan2019federated,collins2021exploiting,oh2021fedbabu} split the local model into a shared part to be aggregated with those from other clients, and a private part of maintaining locality. The shared part is used to transfer public knowledge among clients, and the private part is used to adapt to local data distribution. Clustered FL~\cite{dinh2021fedu} groups clients according to the similarity of the local parameters and trains an aggregated model for each group of clients. Clustered FL extracts common knowledge from similar clients within a group to generate a unified model for the group. These methods, however, still fall short in two aspects. First, these approaches rely on the effectiveness of aggregating or clustering the shared parts and may fail with highly heterogeneous data. Second, these methods simply extract the common knowledge and implement the personalization at the model level, while ignoring the potential at the data level, which may further strengthen the personalized adaptation between the aggregated model and local dataset.


In the community of computer vision~(CV), both visual prompts~\cite{liu2021pre} and adversarial reprogramming~\cite{elsayed2018adversarial} employ a set of learnable parameters as a continuous task-specific vector, which can be tuned based on training data from the downstream task. Visual prompts can effectively help a large-scale pre-trained model achieve fast task transfer by simply training task-related prompts without changing any pre-trained model parameters. The prompt parameters are like the attention guidance to implicitly hint at the task-related information for improving model performance on the new task~\cite{liu2021pre}. This motivates us to regard the different clients as different tasks and adopt client-specific prompts to fine-tune the aggregated model on each local client, which helps to incorporate the extracted global features with the personalized ones.

Based on this insight, we propose a novel PFL framework named pFedPT. Our approach addresses the shortcomings mentioned above by using a visual prompt to implicitly provide a hint of the data distribution on a client for the aggregated model locally. Specifically, each client model integrates a learnable \textit{Prompt Generator} with a backbone participating in aggregation for classification. The prompt generator is a set of locally learnable parameters that can generate personalized visual prompts for its affiliated clients based on their local data distribution. During local training of pFedPT, the generated personalized visual prompts are added to the images. Fig.~\ref{fig:dif} (a) and (b) show the difference in the training process between FedAvg and pFedPT. For different classes of data, Fig.~\ref{fig:dif} (c) and (d) show that the generated prompts increases the inter-class variation (Inter-Var) while decreasing the intra-class variation (Intra-Var). Different class data with a visual prompt is easily distinguished by an aggregated backbone, thereby improving the local performance of the local clients. Then, the backbone is trained on the input composed of raw data and visual prompts to learn the distribution information contained in the prompt. Upon achieving convergence of the two models through alternate training, the backbone implements the extraction of common knowledge from clients and can recognize the visual prompts of different clients. The generated visual prompt reflects the client's characteristics as a client-conditional vector and implements fine-tuning of the backbone in the local client. As a result,  the backbone can capture implicit knowledge about the client's data distribution based on the visual prompt and therefore obtain a better-personalized model. On the other hand, the visual prompt can be of independent interest and added as a plugin for other FL algorithms. It can fine-tune the model received by clients, which can implement the personalized improvement of the model trained by FL algorithms in different clients or further boost the performance of PFL algorithms.

We validate pFedPT on two image classification datasets, including CIFAR10~\cite{krizhevsky2009learning} and CIFAR100~\cite{krizhevsky2009learning}. Empirical results show that pFedPT beats other SOTA baselines of PFL with a 1\%-3\% improvement in test accuracy. In summary, our main contributions are four-fold:
\begin{itemize}
    \item To the best of our knowledge, this is the first work that proposes to use client-specific prompts to help the aggregated models achieve better local adaptation and generalization by leveraging the personalized features of clients. 

    \item We propose a novel PFL framework, dubbed pFedPT, for federated image classification tasks that use the visual prompts from each client to fine-tune the aggregated model and imbue the aggregated local model with information about the local data distribution. 

     \item We show that pFedPT can integrate with several existing FL and decoupled PFL methods to boost their performance, which may be of independent interest. 
 
    \item We conduct extensive experiments to evaluate the effectiveness of pFedPT, which significantly outperforms several SOTA baselines on CIFAR10 and CIFAR100 datasets. Besides, the experimental results illustrate that the prompt can indeed learn personalized knowledge related to the client.
\end{itemize}

\section{Related Work}
\begin{figure*}
    \centering
    \includegraphics[width=0.9\textwidth]{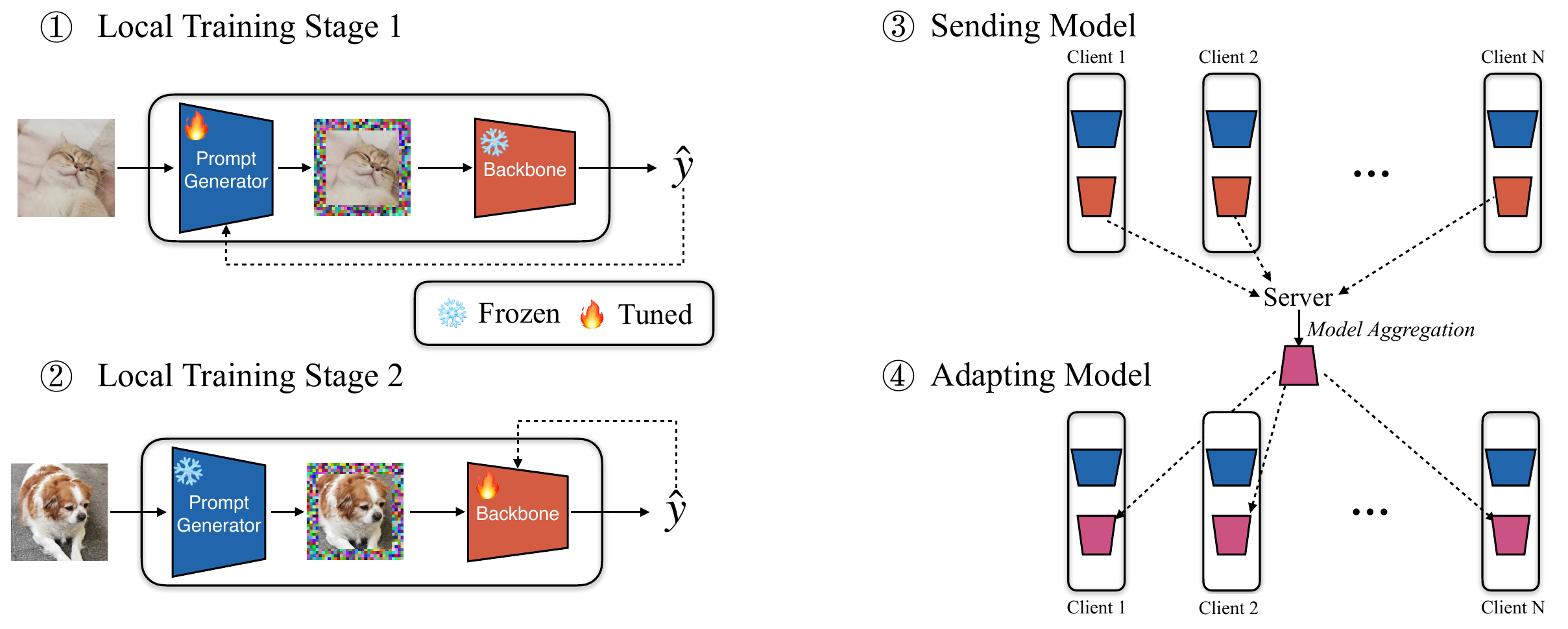}
    \vspace{-0.8em}
    \captionsetup{font=small}
    \caption{The pipeline of the pFedPT. $\hat{y}$ stands for the predicted logits of all classes. The dashed lines in steps 1 and 2 represent the loss backward for the model update. Each client contains a Prompt Generator, a set of personalized learnable parameters preserved locally, and a Backbone, which the server will aggregate with those of other clients. The raw image input will be added to a visual prompt (colored pixels padded around the image) and then passed into the backbone for prediction.}
    \label{fig:pipeline}
\end{figure*}

\paragraph{Personalized Federated Learning (PFL).}\ 
 PFL has drawn significant research interests~\cite{tan2022towards,fallah2020personalized,t2020personalized,hanzely2021personalized,cho2021personalized,wang2023fedabc}. The main difficulty of PFL is to characterize the data distributions of clients and integrate them into the federated learning training process, followed by providing a personalized local model for each client. Currently, the core idea of PFL is to decouple the model into shared layers for feature extraction and personalized layers for classification~\cite{arivazhagan2019federated,collins2021exploiting,oh2021fedbabu}. Each client's parameters of the shared layer are generally updated globally using the FedAvg~\cite{mcmahan2017communication} algorithm. In contrast, the personalized layers are trained locally and will not be shared with others. Those works focus on training a general feature extractor and a personalized classifier head for personalization. 

Other work aims to combine other related machine-learning techniques with PFL. Briggs \etal ~\cite{briggs2020federated} and Mansour \etal ~\cite{mansour2020three} use the clustering technique to divide similar clients into groups and learn a separate model for each group without inter-group federation. Smith \etal ~\cite{dinh2021fedu} use multitasking learning to take advantage of shared representations between clients to improve the generalization performance of each model. Yang \etal ~\cite{yang2020fedsteg} and Chen \etal ~\cite{chen2020fedhealth} use transfer learning to enhance local models by transferring knowledge between relevant clients. T Dinh \etal ~\cite{t2020personalized} add regularizers to the aggregated model to prevent customers' models from simply overfitting their own data sets. Chen \etal ~\cite{chen2018federated} and Fallah \etal~\cite{fallah2020personalized} attempt to develop a well-initialized shared aggregated model using a model-agnostic meta-learning (MAML) approach~\cite{finn2017model}. In addition, fine-tuning using the aggregated model learned by the FedAvg algorithm can also improve the performance of personalized local models~\cite{jiang2019improving,huang2023fusion}. The previous works enable the model to recognize the characteristics of clients and implement personalization for them. However, adding additional information at the data level to achieve better backbone performance on the client has always been ignored. Our pFedPT framework uses visual prompts to implicitly represent the data distribution of clients, which achieves personalization by incorporating the characteristics of clients into the training process at the data level.

\paragraph{Prompt Learning.}\ 
Prompt learning~\cite{liu2021pre}, as a novel application paradigm for large-scale pre-trained models, was first proposed in NLP, and refers to prepending a language instruction to the original text input~\cite{li2021prefix}. In this way, pre-trained models can be given hints about what tasks are currently being performed, thereby achieving strong generalization to downstream transfer learning tasks without fine-tuning the whole model~\cite{floridi2020gpt}. Compared to hard prompts, soft prompts avoid the trouble of manual design, and are more expressive. Lester \etal~\cite{lester2021power} use task-specific continuous vectors as soft prompts and can be optimized by training. In the CV area, Radford \etal~\cite{radford2021learning} propose the CLIP model using language prompts to solve the vision-language tasks, which is similar to the following works~\cite{tsimpoukelli2021multimodal,yao2021cpt}. In~\cite{bahng2022exploring}, the visual prompts are designed as an input-agnostic perturbation, which is padded around the input images. The perturbation-generating function includes a small number of trainable parameters, which helps the pre-trained vision models perform downstream tasks without fine-tuning any parameters. Visual Prompt Tuning (VPT)~\cite{jia2022visual} is introduced as a parameter-efficient alternative to full fine-tuning for pre-trained model~\cite{dosovitskiy2020image}. 

Notably, two concurrent works, PROMPTFL~\cite{guo2022promptfl} and FedPrompt ~\cite{zhao2022reduce} also introduce prompt learning into FL. However, several significant differences exist between our work and these two works. (i) Different training objectives: the goal of PROMPTFL and FedPrompt is to fine-tune existing pre-trained models in the FL system. While pFedPT implements model training from scratch for achieving PFL. (ii) Different ways of training: PROMPTFL and FedPrompt freeze the pre-trained model during training and share the parameter information of the prompt. In pFedPT, backbone and prompt generators are trained alternately during training, and each client has its own unique prompt after training to achieve the goal of personalization. (iii) Different verification scenarios: PROMPTFL and FedPrompt are tested on NLP datasets, while pFedPT is mainly concerned with the improvement brought by the prompt in the CV domain.

\section{Methodology}\label{methodology}
In this section, we introduce the proposed visual prompt based personalized federated learning (pFedPT) framework. Below, we first provide several preliminaries on PFL.

\subsection{Problem setup}
Suppose that there are $N$ clients, denoted as $C_1, ..., C_N$,  respectively. Client $C_i$ has a local dataset $\D^i$. The goal of traditional FL~\cite{mcmahan2017communication} is to collaboratively learn a machine learning model $w$ over the dataset $\D\triangleq \bigcup_{i\in[N]}\D^i$ with the help of a central server, while the raw data are not exchanged. The objective of FL is defined below:
\begin{equation}\label{FL-formulation}
    \argmin_{w} \mathcal{L}(w) = \sum_{i=1}^N \frac{|\D^i|}{|\D|}L_i(w),
\end{equation}
where $L_i(w) = \mathbb{E}_{(x,y)\sim \D^i} [\ell_i(w; (x, y))]$ is the empirical loss of $C_i$.  However, rather than aiming at a single aggregated model w in FL, PFL is supposed to train personalized models $w_i$ for different clients~\cite{tan2022towards}, which is defined as the following optimization problem:
\begin{equation}
    \argmin_{{W}} \mathcal{L}({W}) = \sum_{i=1}^N \frac{|\D^i|}{|\D|}L_i(w_i),
    \label{eq2}
\end{equation}
where $ W = \left\{w_1, ..., w_N \right\}$ is the personalized models set for all clients.

\subsection{Workflow of pFedPT}
We introduce a novel visual prompt based PFL framework for solving the PFL task, dubbed pFedPT. The central insight of the pFedPT is to train the learnable continuous visual prompts about data distribution for each client and use them to fine-tune backbones locally on those clients. Prompts on each client can serve as prior knowledge aiding the backbone to complete the training task. The pFedPT currently focuses on visual-related tasks, wherein each client maintains a prompt generator and a backbone, as shown in Fig.~\ref{fig:pipeline}. When performing image classification tasks, pFedPT first adds prompts generated by the local prompt generator to each image, which is then passed into the backbone for classification prediction. Generally, a complete pFedPT training process mainly includes four steps, as shown in Fig.~\ref{fig:pipeline}:
\begin{itemize}
\item \textbf{Step 1.}\ To begin with, the parameters of the prompt generator on each client are updated with local data while the whole backbone is frozen. 

\item  \textbf{Step 2.}\ After training several epochs, the prompt generator will be frozen, and the backbone will begin to update for a fixed number of epochs.

\item  \textbf{Step 3.}\ When the training process of all clients is finished, they send the trained backbone to the server, followed by the aggregation operation conducted by the server.

\item  \textbf{Step 4.}\ The aggregated backbone will be broadcast to every client to replace the old backbone stored locally. 
\end{itemize}
Repeat the \textbf{Step 1}-\textbf{Step 4}  until the training process of the prompt generator and backbone converges. At this point, the prompt generator generates prompts for each client based on local data distribution and can be seen as a guide to fine-tuning the prediction results of the backbone for the input images. Since the prompt is client-specific, the same backbone can generate different fine-tuning effects when used by different clients to achieve personalization. 

Below, we specify the key components of pFedPT, i.e., prompt generator, which is parameterized with the parameter $\delta$ for the prompt. The prompt is added to the input image to form a prompted image $X_i + \delta_i$. During the local evaluation, the optimized prompt is added to all test images,
\begin{equation} \label{f4}
    \mathcal{X}_i = \left\{x_i^1 + \delta_i, \ldots , x_i^n + \delta_i \right\}.
\end{equation}
There are several ways to design a visual prompt in terms of template and size. Following the settings of ~\cite{bahng2022exploring}, we explore three visual templates: pixel patch at a random location, pixel patch at a fixed location, and padding. We explore various prompt sizes $p$, where the actual number of parameters is $Cp^2$ for patches and $2Cp(H + W - 2p)$ for padding, where $C, H, W $ are the image channels, height, and width, respectively. In order to explore the effect of different prompts on the results, we conducted an experiment on CIFAR10 dataset with a Dirichlet ($0.3$) partition. Fig.~\ref{fig:p_type} shows that padding prompts with $p=4$ size achieve the best performance over other design choices. We use this as the default for all our experiments.

\subsection{Modeling for pFedPT}
Our goal is to learn a personalized prompt $\delta_i$ for each client and a backbone $w$. The prompt $\delta_i$ is also trained by the local data. Our objective is to solve the following:
\begin{equation}
    \argmin_{w, \delta_i} \mathcal{L}(w, \delta_i) = \sum_{i=1}^N \frac{|\D^i|}{|\D|}L_i(w, \delta_i),
    \label{eq3}
\end{equation}
where $L_i(w, \delta_i) = \mathbb{E}_{(x,y)\sim \D^i} [\ell_i(w; (x +\delta_i , y))]$ is the empirical loss of $C_i$. To achieve the goal in Eq.~(\ref{eq2}), exiting PFL algorithms usually add a regularizer to the model to perform information exchange between clients~\cite{li2021ditto,arivazhagan2019federated}, partition the layers as shared and personalized parts by exchanging the shared layers \cite{tan2022towards}, or interpolate the aggregated model with local models \cite{li2021ditto,mansour2020three}. However, pFedPT still uses the aggregated model $w$ to deliver public knowledge between clients, and personalized knowledge is incorporated by adding $\delta_i$ to the data. Specifically, the shared backbone is responsible for the extraction of the common knowledge of each client and identifying the information carried by the visual prompt of the individual clients. The client-specific prompt is responsible for increasing the guidance of the backbone to achieve fine-tuning to adapt to the client's data distribution. We implement personalized prediction of the backbone at the client data level.


\begin{algorithm}[t]
\small
\SetNoFillComment
\LinesNumbered
\SetArgSty{textnormal}
\KwIn{number of communication rounds $T$, the set  of clients  $ \left\{C_1, ..., C_N\right\}$, number of local epochs $E_b$ for backbone, number of local epochs $E_g$ for the prompt generator, learning rate $\eta_b$ for backbone, learning rate $\eta_g$ for the prompt generator, initialization parameters $w^0$ for backbone, initialization parameters $\delta_i^0$ for the prompt generator in client $i$. }
\KwOut{The final model $w^T$}
\BlankLine
\textbf{Server executes}:
initialize $w^0$\\
 $\mathcal{S:}$ choose a random set of devices from $C$ \\
\For {$t=0, 1, ..., T-1$}{
    \For {$C_i  \in \mathcal{S}$ \textbf{in parallel}}{
        send the aggregated model $w^t$ to $C_i$
        
        $w_i^t \leftarrow$ \textbf{LocalTraining}($i$, $w^t$)
    }
    $w^{t+1} \leftarrow \sum_{k=1}^{\left\|\mathcal{S}\right\|} \frac{|\D^i|}{|\D|} w_k^{t}$
}
return $w^T$
\BlankLine
\textbf{LocalTraining}($i$, $w^t$):
$w_i^t \leftarrow w^t$

\For{epoch $i = 1, 2, ..., E_g$}{
    \For{each batch $\textbf{b} = \{x, y\}$ of $\D^i$}{
    \textit{Training for prompt generator:}
        $\delta_i^t \leftarrow \delta_i^t - \eta_g \nabla \ell_i(w_i^t; (x +\delta_i , y)))$
    }
}
\For{epoch $i = 1, 2, ..., E_c$}{
    \For{each batch $\textbf{b} = \{x, y\}$ of $\D^i$}{
        \textit{local backbone training:}
        $w_i^t \leftarrow w_i^t - \eta_b \nabla \ell_i(w_i^t; (x +\delta_i , y))$ 
  }
}
return $w_i^t$ to server
\caption{\ pFedPT framework}
\label{alg:pFedPT}
\end{algorithm}

\subsection{Optimization for pFedPT}

 To achieve the optimization goal of Eq.~(\ref{eq3}), we alternately update the prompt generator and the backbone on each client using gradient descent. pFedPT first trains the prompt generator with the aggregation model fixed, and the model maximizes the likelihood of the correct label y, which is equivalent to solving: 
\begin{equation}
    \argmin_{ \delta_i} \mathcal{L}_i(w, \delta_i) =  \mathbb{E}_{(x,y)\sim \D^i} [\ell_i(w; (x +\delta_i , y))].
\end{equation}
After updating the prompt generator locally, we freeze the parameters of the prompt generator, and then train the backbone for several epochs. The backbone has the following objective function in the client $i$ during the training process:
\begin{equation}
    \argmin_{ w} \mathcal{L}_i(w, \delta_i) =  \mathbb{E}_{(x,y)\sim \D^i} [\ell_i(w; (x +\delta_i , y))].
\end{equation}
A locally trained backbone can learn the client data distribution corresponding to a prompt on the client and prompt knowledge is passed between clients via model aggregation at the server. The backbone on the server aggregates according to the following formula:
\begin{equation}
    w^{t+1} \leftarrow \sum_{k=1}^N \frac{|\D^i|}{|\D|} w_k^{t},
\end{equation}
where $t$ represents the number of training rounds. We summarize the detailed  procedures of pFedPT in Algorithm \ref{alg:pFedPT}.

\begin{figure}
    \centering
    \includegraphics[width=0.5\textwidth]{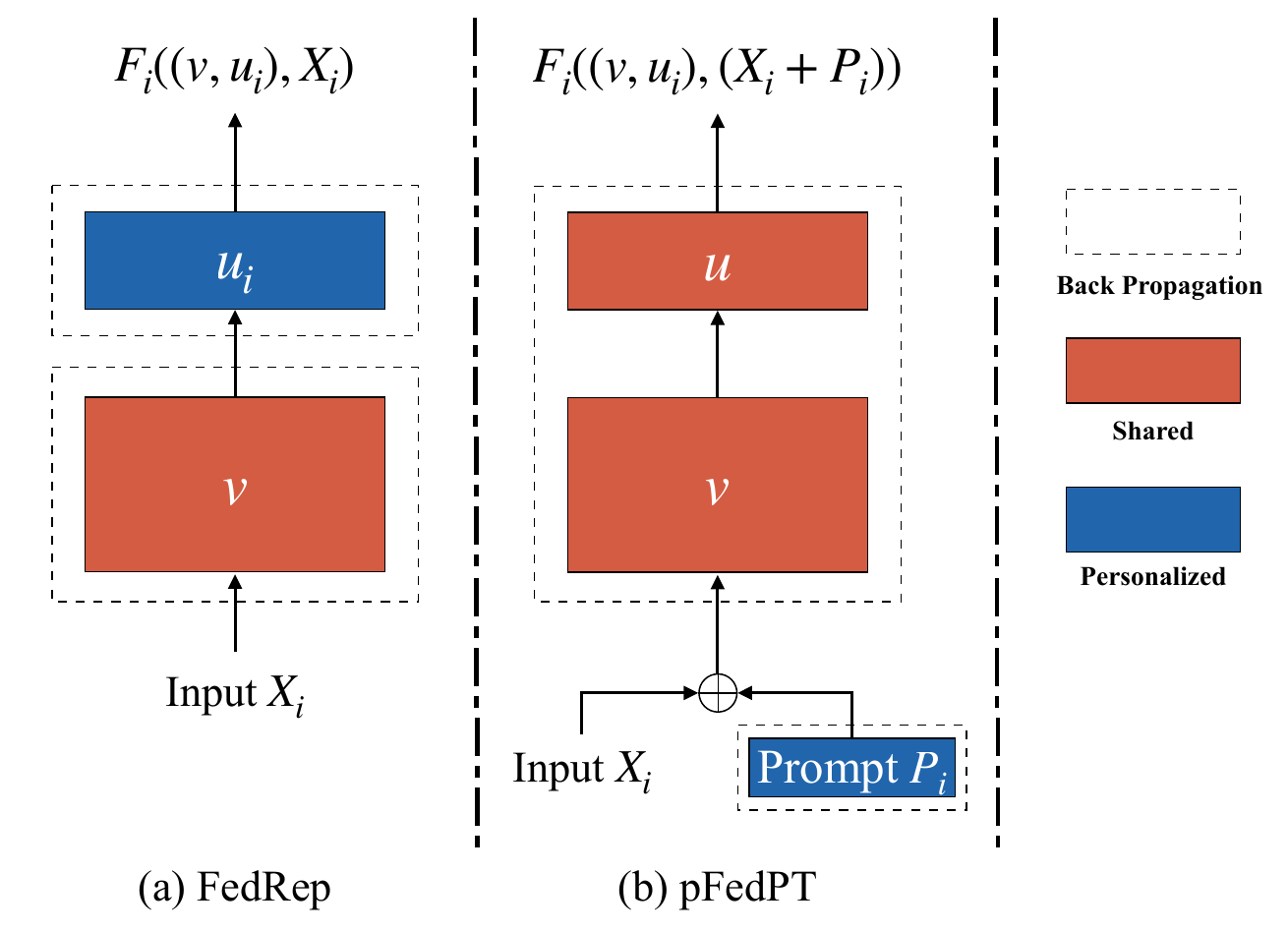}
    \vspace{-1.8em}
    \captionsetup{font=small}
    \caption{Differences between  pFedPT and decoupled personalized FL algorithm (FedRep)}
    \label{fig:dirrp}
\end{figure}

In the end, we give several comments on the differences between our pFedPT and decoupled FedRep \cite{collins2021exploiting}. Fig.\ref{fig:dirrp} describes their training process. Note that pFedPT also has a private part and a public part, but the private part is the prompt generator that we added at the client data level additionally. The personalized visual prompt generated adds the client's personalized knowledge to the training process by fine-tuning backbone's input without changing backbone's inference process. FedRep is to separate the private part in the inference model, and different clients have different inference processes. The objective functions of FedRep and pFedPT are also different. Furthermore, the visual prompt is orthogonal to FedRep type methods, which can be integrated together to further boost their performance.

\section{Experiments}
In this section, we evaluate the effectiveness of pFedPT and compare it with several advanced methods in various datasets and settings. We also conduct a number of exploratory experiments to find out how pFedPT works and verify the effectiveness of pFedPT in terms of  client data distribution. The detailed experimental setup and the more experimental results can be found in the Appendix.

\begin{table*}[!ht]
\centering
\captionsetup{font=small}
\caption{The results of pFedPT and baseline methods on the image datasets with different non-IID settings}
\vspace{-0.8em}
\label{results}
\renewcommand\arraystretch{1.25}
\resizebox{0.9\textwidth}{!}{
\begin{tabular}{lccccccccccccc}
\hline
\textbf{} & \multicolumn{6}{c}{CIFAR10} & \multicolumn{6}{c}{CIFAR100} &\\ \cline{2-13} 
\#setting & \multicolumn{2}{c}{IID} & \multicolumn{2}{c}{Dirichlet } & \multicolumn{2}{c}{Pathological} & \multicolumn{2}{c}{IID} & \multicolumn{2}{c}{Dirichlet } & \multicolumn{2}{c}{Pathological}  \\ \cline{2-13} 
\#client & ViT & CNN & ViT & CNN & ViT & CNN & ViT & CNN & ViT & CNN & ViT & CNN \\ \hline

FedAvg  & 60.50 & \textbf{67.13 } & 53.01  & 61.92  & 54.98  & 63.62 & 29.60 & 26.42 & 25.93 & 26.50 & 27.71 & 30.28  \\
FedProx  & 57.04 & 66.94  & 53.14  & 61.95  & 55.02  & 63.29  & 27.71 & 26.29 & 26.00  & 26.48 & 27.84 & 30.52  \\

MOON  & {60.99} & 66.88  & 61.12  & 62.53  & 65.98  & 63.52 & 29.32 & \textbf{26.43} & 24.95 & 26.93 & 27.61 & 29.00  \\


FedPer  & \textbf{61.57} & {51.46 } & {73.16}  & 77.98  & {75.20}  & 79.97  & 29.74 & 10.82 & {36.78} & {27.79} & 35.36 & 31.13  \\

FedRep  & 48.38 & {49.70 } & 74.11  & 77.65 & 74.48  & 78.39  & 17.84 & 9.13 & 35.06 & {27.39} & 36.13 & 32.41  \\
FedMTL  & 45.65 & 45.65  & 68.48 & 73.95  & 65.39   & 70.94 & 17.91  & 7.34 & 26.08 & 25.85 & 25.46 & 26.32  \\
FedBABU  & 50.41 & 61.17  & 74.21  & 80.11 & 74.30  & 80.69 & 20.61 & 22.55 & 36.17 & 31.66 & 35.74 & 35.45  \\
Local  & 45.37 & 39.04  & 68.40  & 73.98  & 64.83  & 70.76  & 18.01 & 7.33 & 26.23 & 25.15 & 24.65 & 25.34  \\
\hline
pFedPT (ours) & 60.01  & 66.09 & \textbf{74.92} & \textbf{80.83} & \textbf{75.42} & \textbf{81.16} & \textbf{31.66} & {26.41} & \textbf{36.80} & \textbf{32.47} & \textbf{36.88} & \textbf{37.98}\\ 

\hline
\end{tabular}
}
\end{table*}

\begin{table}[!ht]
\centering
\captionsetup{font=small}
\caption{The results of baseline methods with prompts on the image datasets with CNN model in Non-IID settings.}
\vspace{-0.8em}
\label{results+pt}
\renewcommand\arraystretch{1.25}
\resizebox{0.45\textwidth}{!}{
\begin{tabular}{lccccc}
\hline
\textbf{} & \multicolumn{2}{c}{CIFAR10} & \multicolumn{2}{c}{CIFAR100} &\\ \cline{2-5} 
\#setting  & \multicolumn{1}{c}{Dirichlet } & \multicolumn{1}{c}{Pathological} &  \multicolumn{1}{c}{Dirichlet } & \multicolumn{1}{c}{Pathological}  \\ \cline{2-5} 
 \hline

FedProx     & 61.95    & 63.29   & 26.48  & 30.52  \\
FedProx+PT    & \textbf{80.47}    & \textbf{81.48}    & \textbf{31.95}  & \textbf{37.88}  \\
\hline
MOON     & 62.53    & 63.52   & 26.93  & 29.00  \\
MOON+PT     & \textbf{77.84}   & \textbf{76.00}    & \textbf{28.67}  & \textbf{34.60}  \\
\hline
FedPer     & 77.98    & 79.97    & {27.79} & 31.13  \\
FedPer+PT     & \textbf{78.40}    & \textbf{80.59}    & \textbf{28.83}  & \textbf{31.14}  \\
\hline
FedRep     & 77.65  & 78.39     & {27.39}  & 32.41  \\
FedRep+PT      & 77.65    & \textbf{79.11 }   & \textbf{29.19}  & \textbf{32.75}  \\
\hline
\end{tabular}
}
\end{table}

\subsection{Experimental Setup}

\paragraph{Baselines.}\ 
We compare the pFedPT with several advanced FL  methods, including FedAvg~\cite{mcmahan2017communication}, 
FedProx~\cite{li2020federated}.
MOON~\cite{li2021model},  FedPer~\cite{arivazhagan2019federated}, 
FedRep~\cite{collins2021exploiting},
FedMTL~\cite{smith2017federated} and FedBABU~\cite{oh2021fedbabu}. We also compare a baseline named \textbf{Local}, where each client trains a model with its local data without federated learning. We conduct experiments on two benchmark datasets:
CIFAR10~\cite{krizhevsky2009learning} and CIFAR100~\cite{krizhevsky2009learning}. CIFAR100 is a more difficult dataset for classification tasks than CIFAR10. We use PyTorch~\cite{oord2018representation} to implement pFedPT and the other baselines.

\paragraph{Datasets.} We consider two different scenarios for simulating non-identical data distributions (Non-IID) across federated clients. Dirichlet Partition follows works~\cite{hsu2019measuring}, where we partition the training data according to a Dirichlet distribution Dir($\alpha$) for each client and generate the corresponding test data for each client following the same distribution. We specify $\alpha$ equal 0.3 for each dataset. In addition, we evaluate with the pathological partition setup similar to \cite{zhang2020personalized}, in which each client is only assigned a limited number of classes at random from the total number of classes. We specify that each client possesses 5 classes for CIFAR10 and 50 classes for CIFAR100. 

\paragraph{Implementation Details.}\ 
We verify the experimental results based on CNN and ViT architectures. The CNN model consists of 2 convolutional layers with 64 $\times$ filters followed by 2 fully connected layers with 394 and 192 neurons and a softmax	layer. We use tiny ViT architecture consisting of 8 blocks with 8 self-attention layers in each block. The corresponding attention head number is 8, the patch size is 4, and the embedding dimension is 128. We set the number of clients to 50, and then each client has a 20\% chance of participating in each communication round. We utilize the SGD algorithm~\cite{cherry1998sgd} as the local optimizer for all methods. We use padding as our prompt method. We set batch size as 16 in the local training phase, the local training epochs for the generator and backbone as 5 in each round, the learning rate for the backbone as 0.005, the learning rate for the prompt generator as 1, and the  padding prompt size as 4. The number of communication rounds is set to 150 for CIFAR10, 300 for CIFAR100, where all FL approaches have very limited or no accuracy gain with more communications.

\subsection{Main Results} \label{main result}
We run vast experiments to determine the superiority of pFedPT on the model performance in different datasets. Our results highlight the benefit of pFedPT compared to the existing PFL optimization approaches.

\paragraph{Better performance of pFedPT.}
Tab.~\ref{results} compares the best accuracy of the pFedPT with baselines on evaluation datasets with various settings. On CIFAR10 and CIFAR100, the pFedPT consistently achieves the best test accuracy with Non-IID setting. For instance, when training on the data of Dirichlet distribution CIFAR10 with CNN, the test accuracy of the pFedPT is 80.83\%, the accuracy of FedAvg is 61.92\%, and the accuracy of the FedPer is 77.98\%.  The improvements of pFedPT indicate that prompts in each client effectively improve the backbone performance in each client. Similarly, in CIFAR100, pFedPT outperforms most baselines in various settings and achieves comparable results in the Dirichlet setting.

\paragraph{Robustness of pFedPT.}  pFedPT achieves clear success on both ViT and CNN models and seems to get better performance as the FL tasks become more difficult (since better performance is observed at a greater Non-IID extent and in datasets that are intrinsically more difficult). Interestingly, in the IID setting, we show that all the personalized solutions exhibit some extent of performance degradation, which becomes more significant as the dataset becomes more challenging. Our interpretation of this phenomenon is that when the data are distributed under the IID setting, the PFL approach does not effectively take advantage of the personalization characteristics among clients, resulting in performance degradation. pFedPT will utilize the data distribution information in the client by visual prompts. When the data is IID, the output will be similar on the various clients and degenerate into the FedAvg.

\paragraph{Improvements  of prompt for other algorithms.}  Visual prompts can improve the performance of backbones on clients by fine-tuning the backbone with hints about the distribution of the client's data. We explore the usefulness of visual prompts as prior knowledge for other FL algorithms, and Tab.~\ref{results+pt} presents these results. In the Dirichlet setting of CIFAR10, the final test accuracy of FedProx increases from 61.95\% to 80.47\% after adding prompts, and the test accuracy of MOON increases from 62.53\% to 77.84\%. We find that a visual prompt enables fine-tuning of the backbone of the client, which helps FL algorithms that pursue high precision fuse client information for personalization. Similarly, PFL algorithms with model decoupling, like FedRep and FedPer, can also yield a performance boost by integrating pFedPT. Therefore, prompt can be used as an additional component to improve the personalization performance of some existing FL algorithms.

Compared with other baselines, the pFedPT takes full advantage of the data improvement space. Additional prompts are added to the data entered into the model to improve the performance of the model on each client.

\subsection{Exploratory Study}
To provide more explanation for pFedPT, we additionally conduct several exploratory studies on pFedPT.



 \begin{figure}[htbp]
    \centering
    \includegraphics[width=0.47\textwidth]{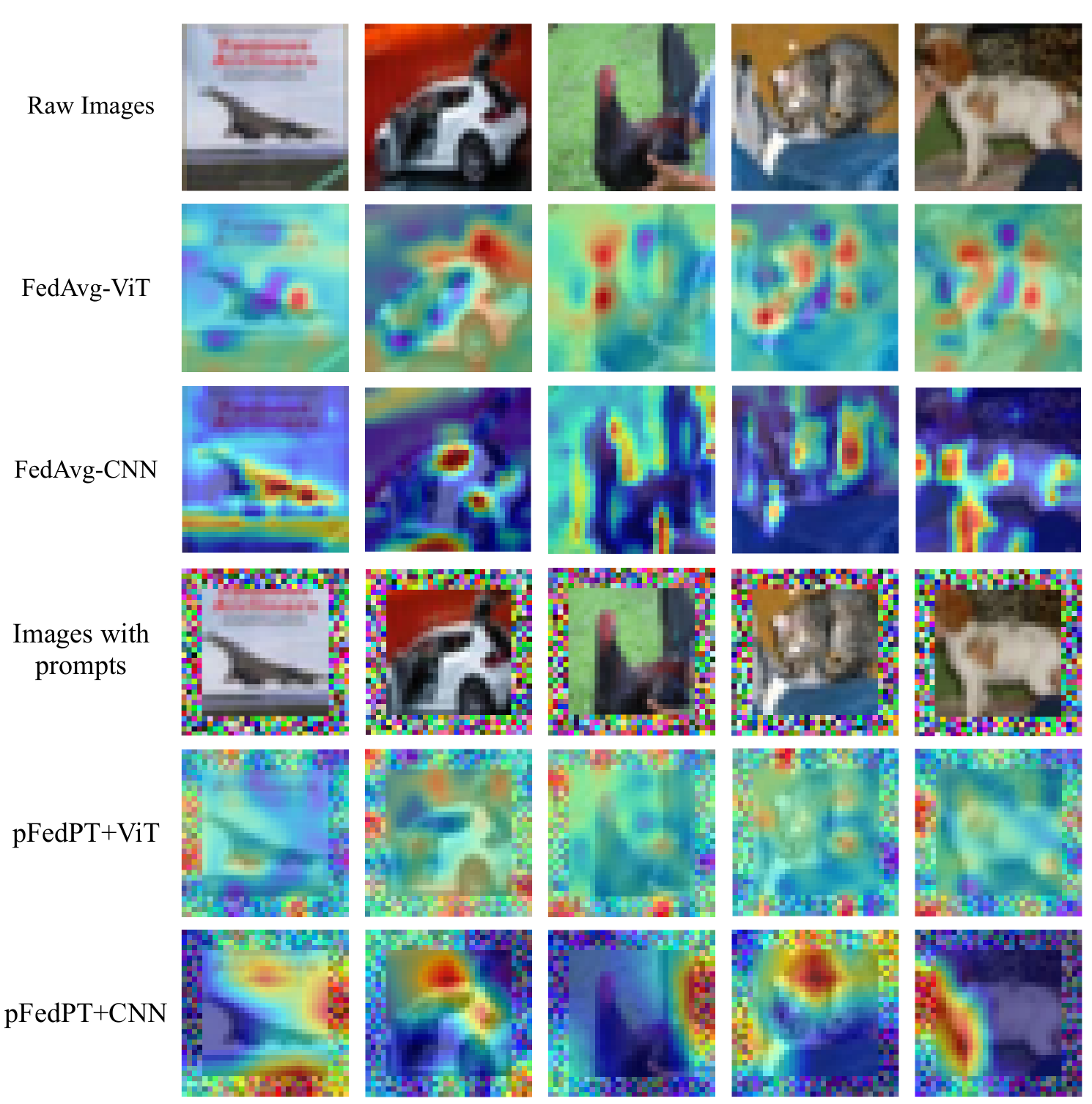}
    \vspace{-0.8em}
    \captionsetup{font=small}
    \caption{Visualization results generated by FedAvg and pFedPT with different backbones.}
    \label{fig:attenmap}
\end{figure}

\paragraph{Visualization of attention maps.} 
To illustrate the effectiveness of visual prompts, we conducted some validation experiments. We train ten clients using FedAvg and pFedPT with ViT and CNN backbones under the Dirichlet setting of the CIFAR10 dataset, respectively. As shown in Fig.~\ref{fig:attenmap}, we make a visualization of the attention map of the last layer in the ViT and CNN by Grad-CAM~\cite{DBLP:journals/ijcv/SelvarajuCDVPB20}.
The first three rows in the figure show that  FedAvg focuses on some salient classification features of the raw image.
The fourth row contains the input images with the padding visual prompts, which are added by the prompt generator of pFedPT according to Eq.~(\ref{f4}). Both pFedPT+ViT and pFedPT+CNN shift some attention to the added prompts, which can help obtain the prior knowledge for the model inference process, thus improving the performance of the model.

 \begin{figure}[htbp]
    \centering
    \includegraphics[width=0.45\textwidth]
    {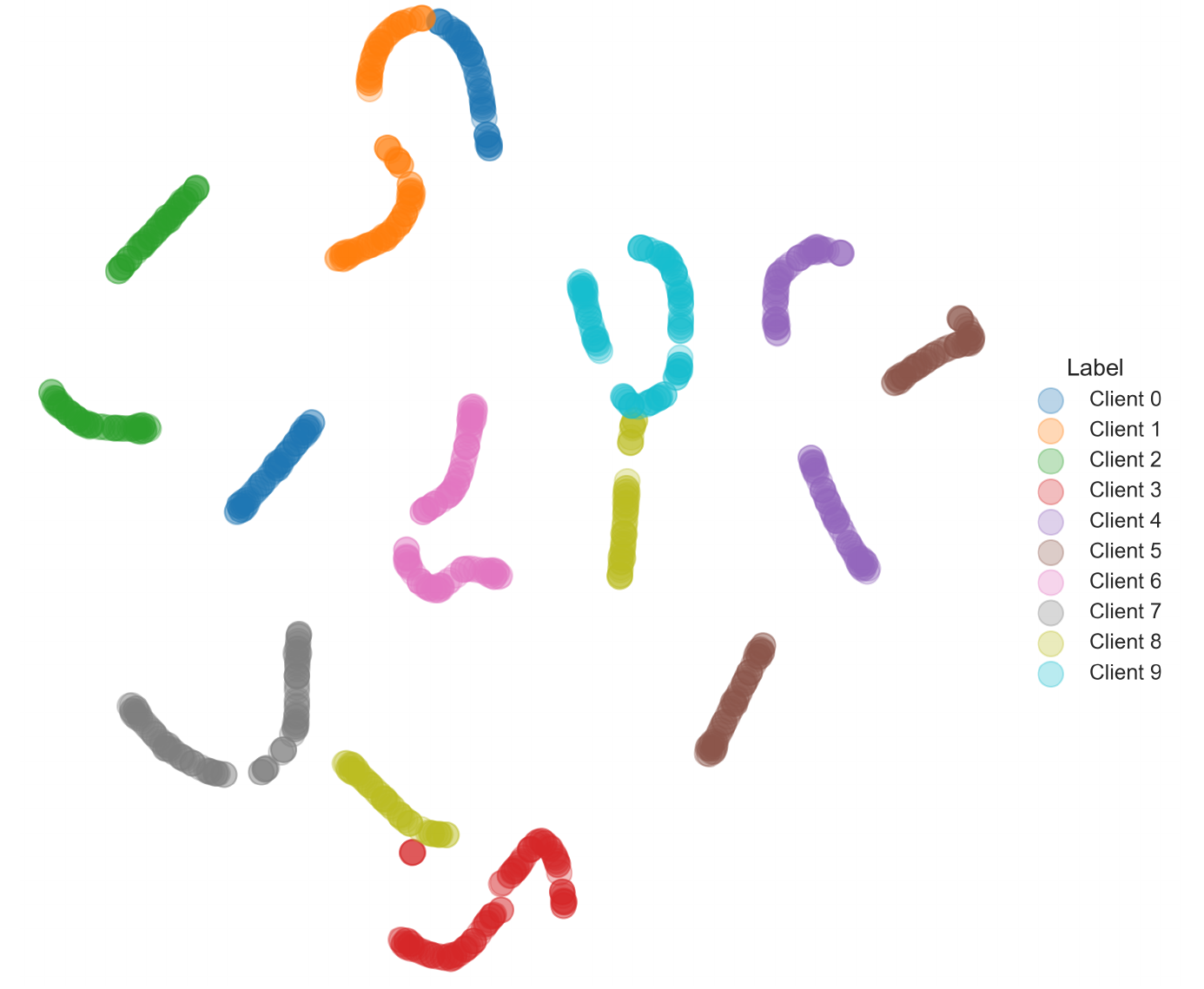}
    \vspace{-0.8em}
    \captionsetup{font=small}
    \caption{t-SNE visualization of embedding for pure color images with learned prompts in different clients.}
    \label{fig:tSNE}
    \vspace{-0.4cm}
\end{figure}

\paragraph{The guidance information contained in the prompts.}\  In order to further explore the influence of visual prompts, we generated 100 different pure color images with the shape of $[3\times32\times32]$. Using the pure color picture, pFedPT can exclude the disturbance of image contents and pay more attention to visual prompts. We feed those color pictures into pFedPT models in different clients with different prompts and visualize the output embeddings of their last MLP layer. We project them into a two-dimensional plane using the t-SNE algorithm~\cite{van2008visualizing}. Fig.~\ref{fig:tSNE} shows that after the visual prompts are added, the model outputs of different clients can be easily distinguished, indicating that the prompts contain prior knowledge of the client model and aid in the classification task. 

\begin{figure}
    \centering
    \includegraphics[width=0.45\textwidth]{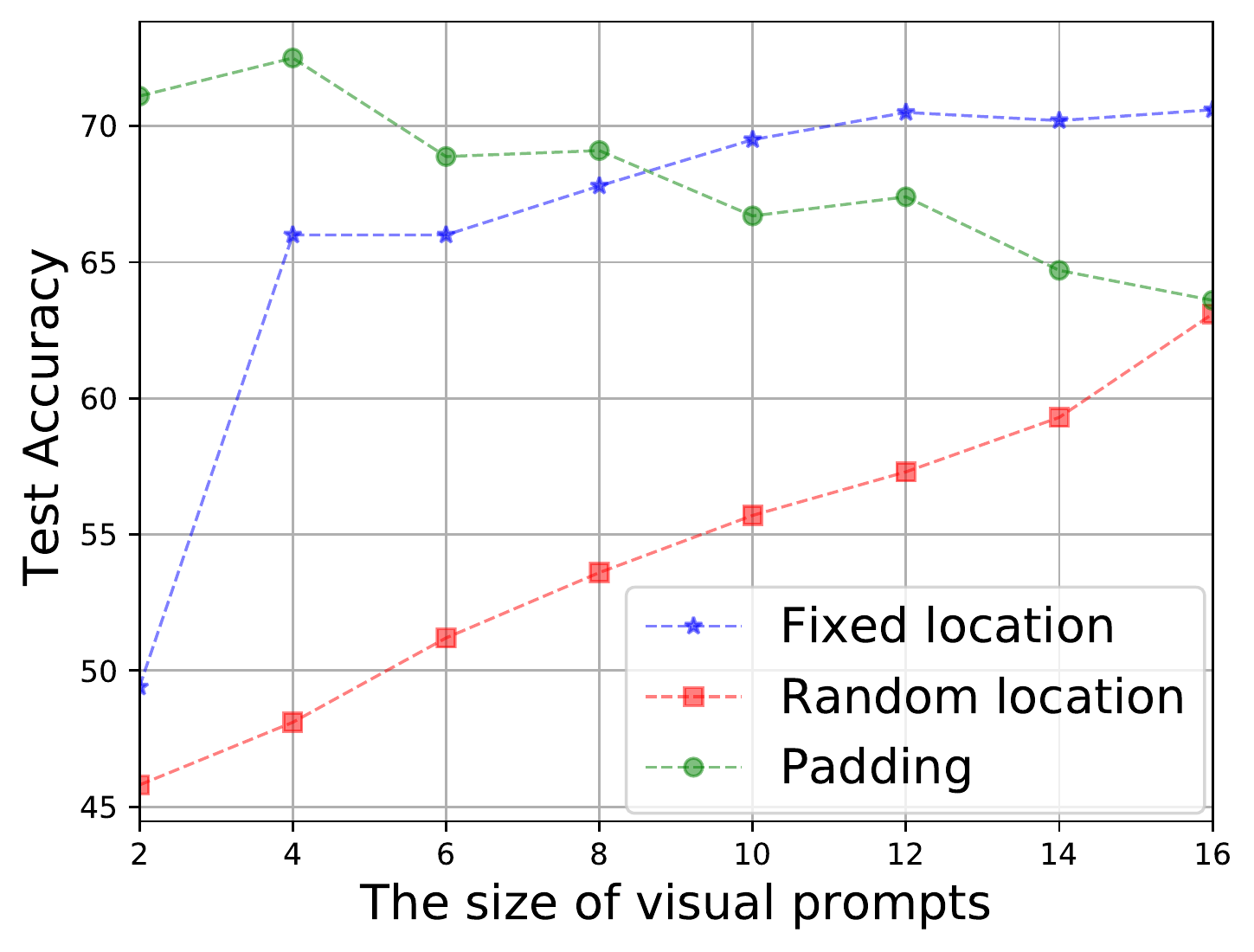}
    \vspace{-0.8em}
    \captionsetup{font=small}
    \caption{Effect of different types of prompts}
    \label{fig:p_type}
\end{figure}

\paragraph{Impact of different types of visual prompts.} 
We analyze different choices on how and where to insert prompts in the input images and how they would affect the final performance. We perform an ablation study on different prompt sizes in $p=\{2,4,6,\dots,16 \}$ in CIFAR10 with a Dirichlet distribution. As shown in Fig.~\ref{fig:p_type}, padding prompts reach the highest performance with a size of 4. The test accuracy of fixed location and random location prompts grows gradually with the increase in prompt size, but it is still slightly lower than the padding prompt. In contrast, the accuracy of padding prompts decreases as the prompt size increases. A possible explanation is that the padding method covers more pixels of the original images than the other two methods when using the same length of prompts. As a result, the key information for classification could be blocked by the prompts and harm the performance of the model. Overall, the padding prompts with size 4 achieve the best performance. Note that other visual tasks may require significantly different kinds of prompts. 

\section{Conclusion}
In this work, we propose a novel framework named pFedPT, a personalized federated learning method based on visual prompts. We make the first attempt to introduce visual prompts to personalized federated learning, using a prompt generator to distill information from local data into the visual prompts and fine-tune the backbone. In the process of pFedPT training, the backbone could use the guidance information from visual prompts to perform the personalized downstream tasks. Since the prompt generator is trained locally on the client, it does not reveal data distribution information about the client to others or the server. pFedPF can also serve a strong plugin to boost the performance of existing FL methods, which could be of independent interest. We provide extensive experiments to illustrate how the pFedPT works and demonstrate its effectiveness in experiments with heterogeneous settings and several types of dataset partition.

{\small
\bibliographystyle{ieee_fullname}
\bibliography{egbib}
}
\appendix
\onecolumn
\section{Appendix: More Experiment Results}
We run experiments on the real-world datasets for image classification tasks, including CIFAR10, CIFAR100, and Tiny ImageNet. We conduct comprehensive investigations for the impact on client heterogeneity by designing IID and non-IID data scenarios. For comparison, we utilize the FedAvg, FedProx, and FedPer algorithms as baselines. The experiment settings are described in detail below. 

\subsection{Setups}

\paragraph{Dataset.}\
We adopt real-world datasets for the image classification task, including CIFAR10, CIFAR100, and Tiny ImageNet. The CIFAR10 dataset contains 50,000 training data and 10,000 test data in 10 classes. Each data sample is a $3\times 32\times 32$ color image. CIFAR100~\cite{krizhevsky2009learning} includes 50,000 training data and 10,000 test data in 100 classes as 500 training samples per class. TinyImageNet~\cite{oord2018representation} involves 100,000 training images and 10,000 test images in 200 classes for $3\times 64\times 64$ color images, as shown in Table~\ref{data}. For CIFAR10/100 and TinyImageNet, we normalize the pixel value within a specific mean and std value in our code, which are [0.5, 0.5, 0.5] for the mean and [0.5, 0.5, 0.5] for the std.
\begin{table*}[ht]	
\captionsetup{font=small}
\vspace{-0.8em}
\caption{The similarity between predicted and real data distribution}
	\label{data}
	\centering
	\resizebox{0.6\textwidth}{!}{
		\begin{tabular}{ccccc}
			\toprule
			Datasets & Training Data & Test Data  & Class & Size \\
			\midrule
			CIFAR-10 & 50,000 & 10,000 & 10 & $3\times32\times32$ \\
			CIFAR-100 & 50,000 & 10,000 & 100 & $3\times32\times32$ \\
                TinyImageNet & 100,000 & 10,000 & 200 & $3\times64\times64$ \\
   
			\bottomrule
		\end{tabular}
	}
 \vspace{-1.5em}
\end{table*}

\paragraph{Backbone.}\
We  adopt two backbones including ViT and CNN for experiments.
Given an image $I\in \mathbb{R}^{3\times h\times w}$, the ViT reshapes it to a sequence of flattened 2D patches $I_p \in \mathbb{R}^{n\times (p^2\cdot c)}$, where $c$ is the number of channels. $(h, w)$ is the height and width of the original image, while $(p, p)$ is the size of each image patch. A trainable linear projection flattens the patches into a latent $D$-dimensional embedding space, which is then embedded with a positional embedding. The transformer encoder consists of $N$ layers of Multi-Head Self-Attention (MSA) and Multi-Layer Perceptron (MLP) blocks. For MSA, the queries, keys and values are generated via linear transformations on the inputs for $K$ times with one individual learned weight for each head. Then in parallel, the attention function is applied to all queries, keys, and values. The sequence of image patches $I_p$ is passed into the MSA, followed by MLP for $N$ times to get the final outputs.

The ViT used by us consists of 8 blocks with 8 self-attention layers in each block. The corresponding attention head number is 8, the patch size is 4 and the embedding dimension is 128.
 The CNN  consists of the basic modules of CNN, including two conventional layers with 64 of $5 \times 5$ convolution kernels, each conventional layer followed by a down-pooling larger, after that are two fully connected layers with 394 and 192 neurons and a softmax layer for prediction.

\begin{figure}[h]
    \centering
    \includegraphics[width=0.7\textwidth]{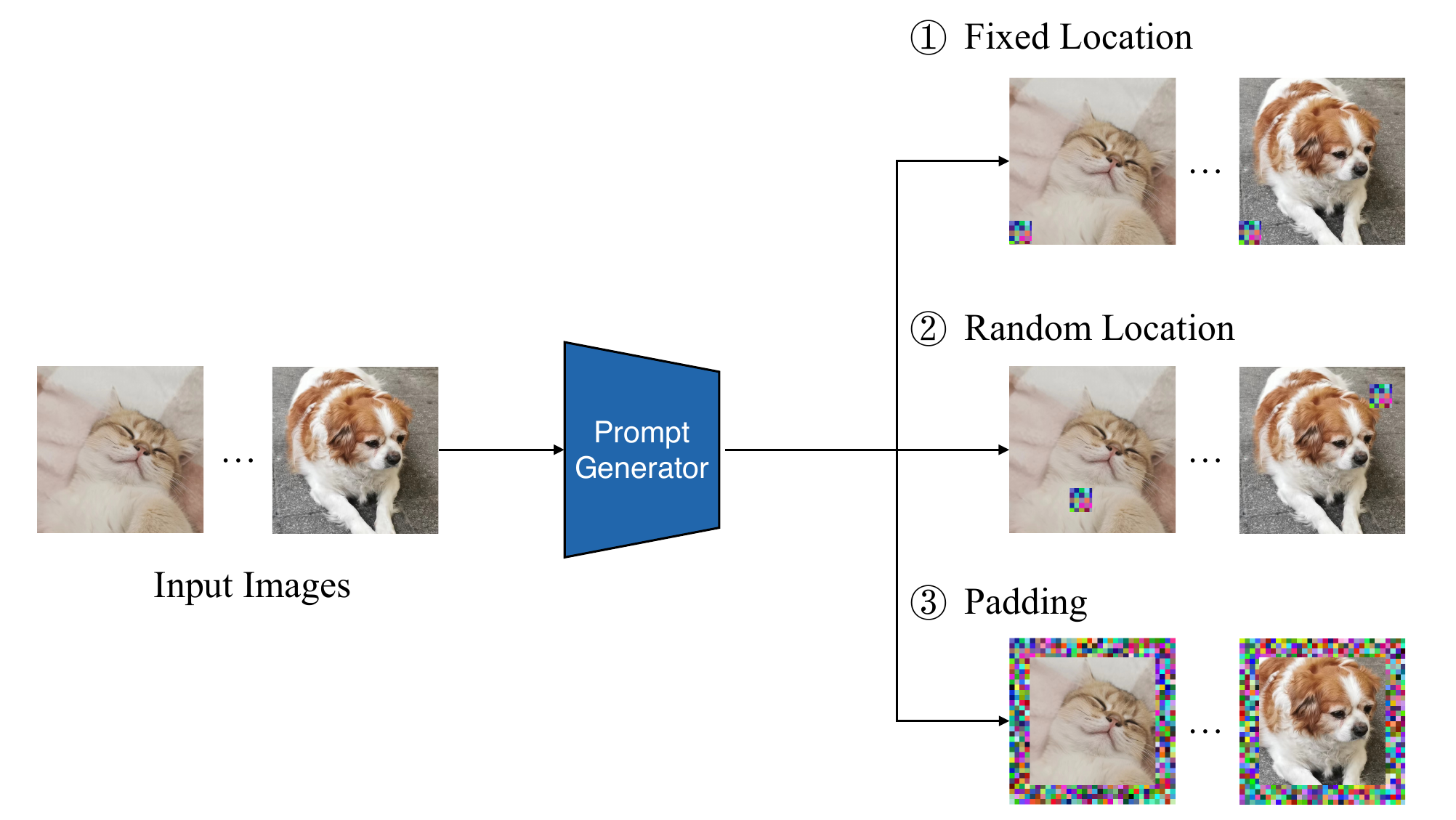}
    \captionsetup{font=small}
    \vspace{-0.8em}
    \caption{Different visual prompt templates}
    \label{fig:templates}
    \vspace{-1.5em}
\end{figure}

\paragraph{Three types of generating Visual Prompts.}\
Following the settings of ~\cite{bahng2022exploring}, we explore three visual templates: pixel patch at a random location, pixel patch at a fixed location, and padding. We describe in Fig.~\ref{fig:templates} the difference between different visual prompts and how these visual prompts templates are added to the input picture. The comparison results are shown as Fig.~\ref{fig:p_type}.

\begin{figure*}[h]
\centering
\subfloat[IID]{\includegraphics[width=0.32\textwidth]{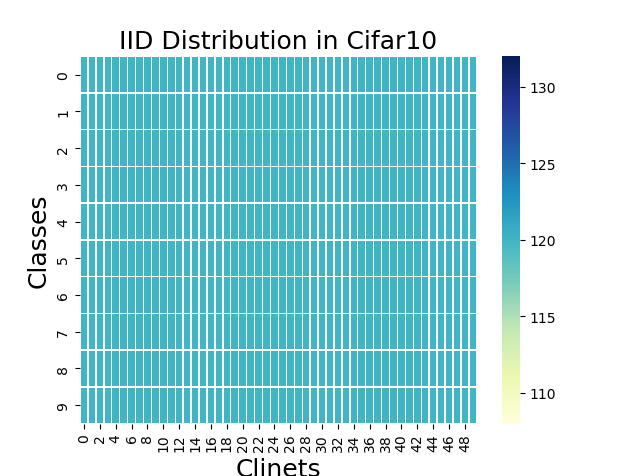}%
}
\hfill
\subfloat[Dirichlet]{\includegraphics[width=0.32\textwidth]{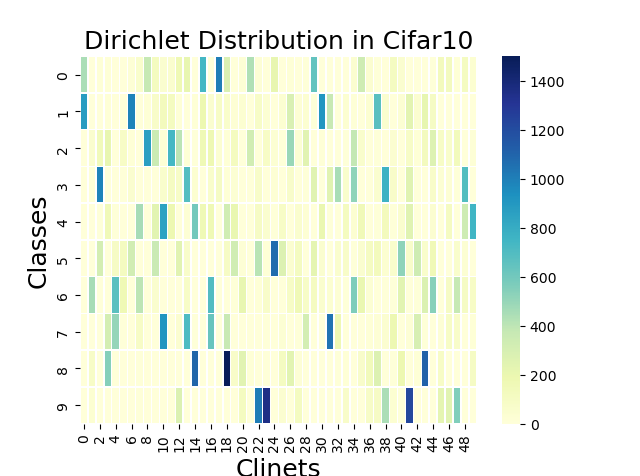}%
}
\hfill
\subfloat[Pathological]{\includegraphics[width=0.32\textwidth]{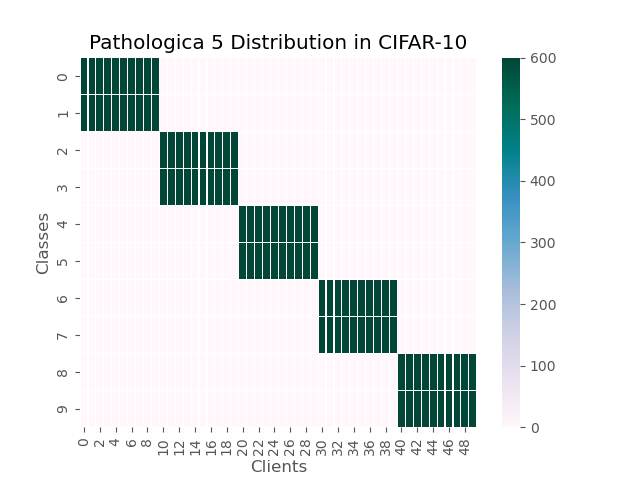}}
\captionsetup{font=small}
\vspace{-0.8em}
\caption{Heat maps for each client with  CIFAR10  dataset under different data partitions. The color bar denotes the number of data samples. Each rectangle represents the number of data samples of a specific class in a party. }
\label{fig:datadis}
\vspace{-1.5em}
\end{figure*}

\paragraph{Dataset Partitions.}\
To fairly compare with the other baselines, we introduce heterogeneity by splitting the total dataset and sampling the label ratios from the Dirichlet distribution and Pathological distribution. An additional parameter is used to control the level of heterogeneity of the entire data partition. In order to visualize the distribution of heterogeneous data, we make heat maps of the label distribution in different datasets, as shown in Fig.~\ref{fig:datadis}.  It could be seen that for heterogeneity weight equal to 0.3 in the Dirichlet distribution, about 10\% to 20\% of the categories dominate on each client, which is the blue block in Fig.~\ref{fig:datadis}. For heterogeneity weight equal to 5 in pathological distribution, 50\% of the categories dominate on each client, which is the black block in  Fig.~\ref{fig:datadis}. The IID dataset is totally averaged for each client, which is the green block in  Fig.~\ref{fig:datadis}.

\paragraph{Baselines.}\
FedAvg~\cite{mcmahan2017communication} is proposed as the basic framework in federated learning. 
FedProx~\cite{li2020federated} adds a proximal term to the objective function of the local model and allows for the emergence of incomplete training of the local model.
MOON~\cite{li2021model} is to utilize the similarity between model representations to correct the local training of individual parties, conducting contrastive learning at the model level. 
FedPer~\cite{arivazhagan2019federated} and  FedRep~\cite{collins2021exploiting} are base + personalization layer approaches for federated training of deep feedforward neural networks, which can combat the ill-effects of statistical heterogeneity.  
FedMTL~\cite{smith2017federated} uses a multi-task learning (MTL) framework to learn separate models for each client. FedBABU~\cite{oh2021fedbabu} achieves good personalization performance by freezing the last discriminative layer of the network and fine-tuning it after training. We also compare a baseline named Local, where each client trains a model with its local data without federated learning.

 \paragraph{Hyper-parameters Selections.}\
We fix the learning rate for local training as 0.005 and for the prompt generator training as 1.0. We fix the training batch size as 16 and fix the epoch for local training as 5. For the specific parameters in FedProx, the proximal rate is set as 0.0001. For the specific parameters in MOON, the $\mu$ is set as 1.0.  For the specific parameters in FedRep, the personalized learning rate is set as 0.01. For the specific parameters in FedMTL, the iterations for solving quadratic sub-problems are set as 4000.  For the specific parameters in FedBABU, the fine tuning step is set as 1. 

\subsection{Other Experimental Results}
\paragraph{Experimental results for TinyImageNet.}
\begin{table*}[h]
\tiny
\centering
\captionsetup{font=small}
\vspace{-0.8em}
\caption{The results of the pFedPT and the benchmark methods on the TinyImageNet dataset with different non-IID settings and backbones.}

\label{results-i}
\renewcommand\arraystretch{1.25}
\resizebox{0.6\textwidth}{!}{
\begin{tabular}{lccccccc}
\hline
\textbf{} & \multicolumn{6}{c}{TinyImageNet}  &\\ \cline{2-7} 
\#setting & \multicolumn{2}{c}{IID} & \multicolumn{2}{c}{Dirichlet } & \multicolumn{2}{c}{Pathological}  \\ \cline{2-7} 
\#client & VIT & CNN & VIT & CNN & VIT & CNN \\ \hline

FedAvg  & 15.47  & 16.22  & 12.06  & 11.01  & 9.37 & 10.53   \\
FedProx  & 15.53  & 15.81  & 13.82  & 10.98  & 12.84 & 11.32   \\
Moon  & 15.20  & \textbf{16.78}  & 15.51  & 11.05  & 13.67 & 10.24   \\
FedPer  & 14.85  & 13.18  & 23.95  & 20.21  & 19.23 & 17.99   \\
FedRep  & 12.91  & 11.56  & \textbf{25.24}  & 21.32  & 23.43 & 20.42  \\
FedMTL  & 9.75  & 11.02  & 21.14  & 17.96  & 18.30  & 17.39  \\
FedBABU  & 13.61  & 12.04  & 24.34  & 18.62  & 22.46 & 18.41  \\

Local  & 9.82  & 10.72  & 20.86 & 17.43  & 18.19 & 17.12 \\
\hline
pFedPT (ours) & \textbf{18.72} & 16.53 & 21.21 & \textbf{21.42} & \textbf{25.95} & \textbf{20.66}\\ \hline
\end{tabular}
}
\end{table*}

We compare the performance of pFedPT and other baselines on the TinyImageNet dataset with 500 rounds of communications. Tab.~\ref{results-i} shows that the experimental results in TinyImageNet are still consistent with the interpretation in Sec.~\ref{main result}. For Pathological distribution with ViT, the test accuracy of the pFedPT is 25.95\%, the accuracy of FedAvg achieves 9.37\% and the accuracy of the FedPer achieves 19.23\%. However, the test accuracy of the pFedPT is not the highest in Dirichlet distribution, our explanation is that it is hard to generate stable prompts in 500 rounds due to the increased task difficulty.
\begin{figure}[h]
    \centering
    \includegraphics[width=0.5\textwidth]{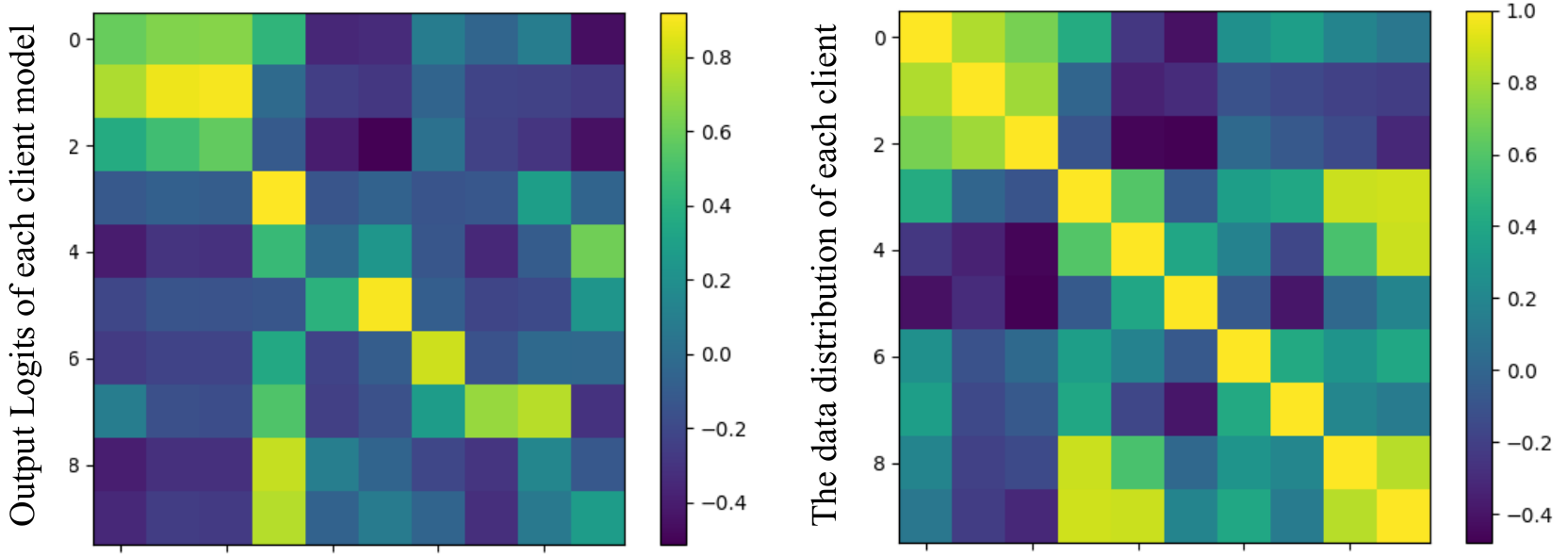}
    \captionsetup{font=small}
    \vspace{-0.8em}
    \caption{Similarity comparison between the distribution of the predicted classes and the distribution of the local data.}
    \label{fig:output_dis}
    \vspace{-1.5em}
\end{figure}
\paragraph{The connection between Prompt and the client data distribution.}
We use a single, pure-color image as input to investigate the relationship between the local model output and the data distribution of each client. Ideally, the output distribution over classes of each pFedPT client should align with the local data distribution. Fig.~\ref{fig:output_dis} reveals that after adding the visual prompts, the outputs of the pFedPT will be similar to the distribution of the client itself. The difference between the visual prompts generated by clients with similar data distribution is also smaller, which means that the visual prompts indeed contain the data distribution information of the clients. Therefore, the visual prompts provide the model with certain prior knowledge when classifying a specific client and assist in the classification task.

\begin{figure}[h]
    \centering
    \includegraphics[width=0.35\textwidth]{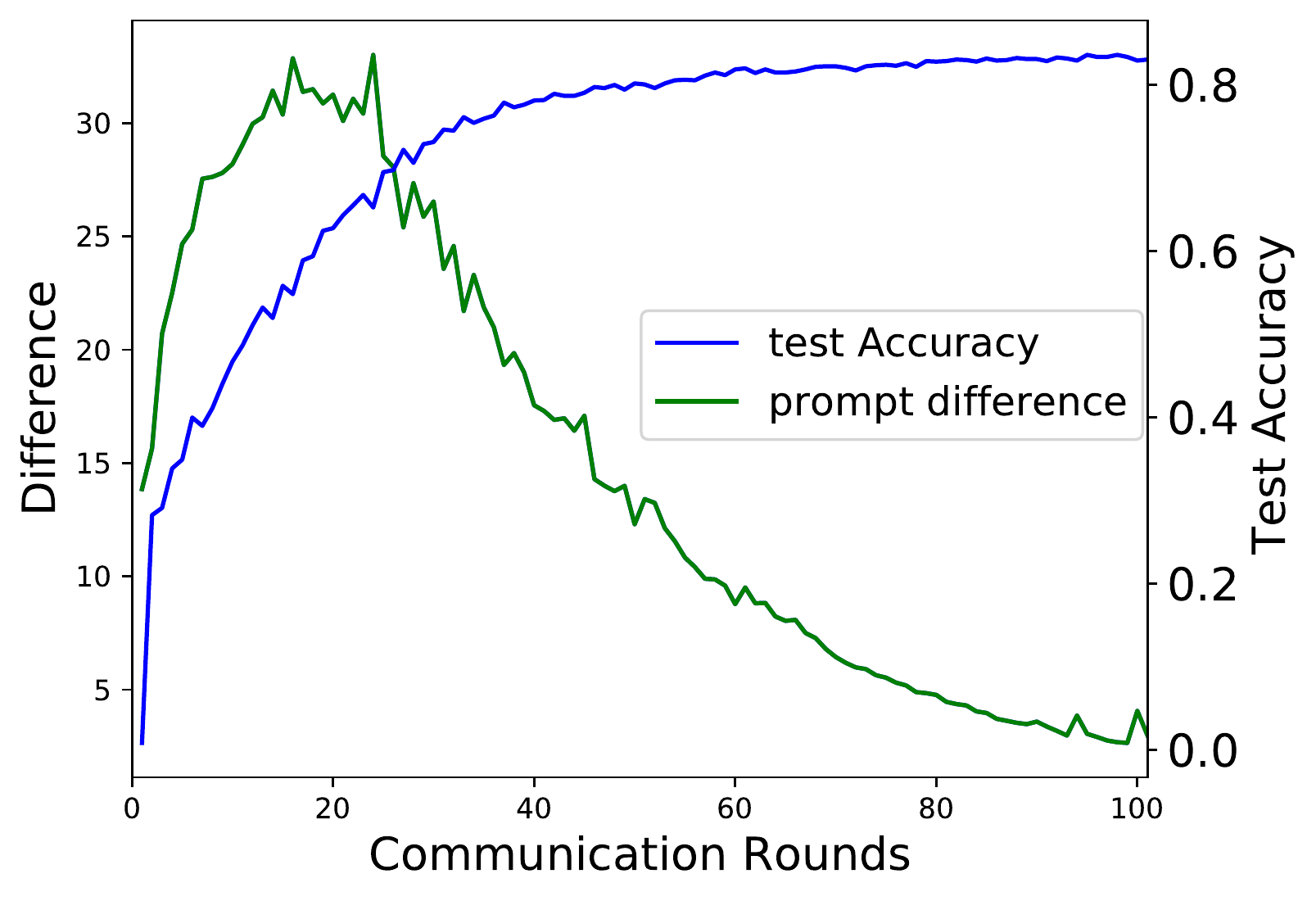}
    \captionsetup{font=small}
    \vspace{-0.8em}
    \caption{The difference of  prompt between two consecutive rounds.}
    \label{fig:difference}
        \vspace{-1.5em}
\end{figure}

\paragraph{Empirical analysis of the learned prompts.}\ 
Fig.~\ref{fig:difference} records the average difference of the prompts generated between the two rounds before and after ten clients during the pFedPT training process. The overall experimental results are divided into two stages: first ascending and then descending. In our settings, the initial prompt generator parameters of each client are the same, and the rising stage is the mapping process between each client and the prompts based on its own data distribution. The descending stage is when the aggregated model tends to converge, and the mapping between the prompt and the client data distribution on each client is complete. Eventually, the change in prompt embedding approaches 0, that is, each client establishes stable prompts that conform to its own data distribution.

\begin{figure}[h]
    \centering
    \includegraphics[width=0.35\textwidth]{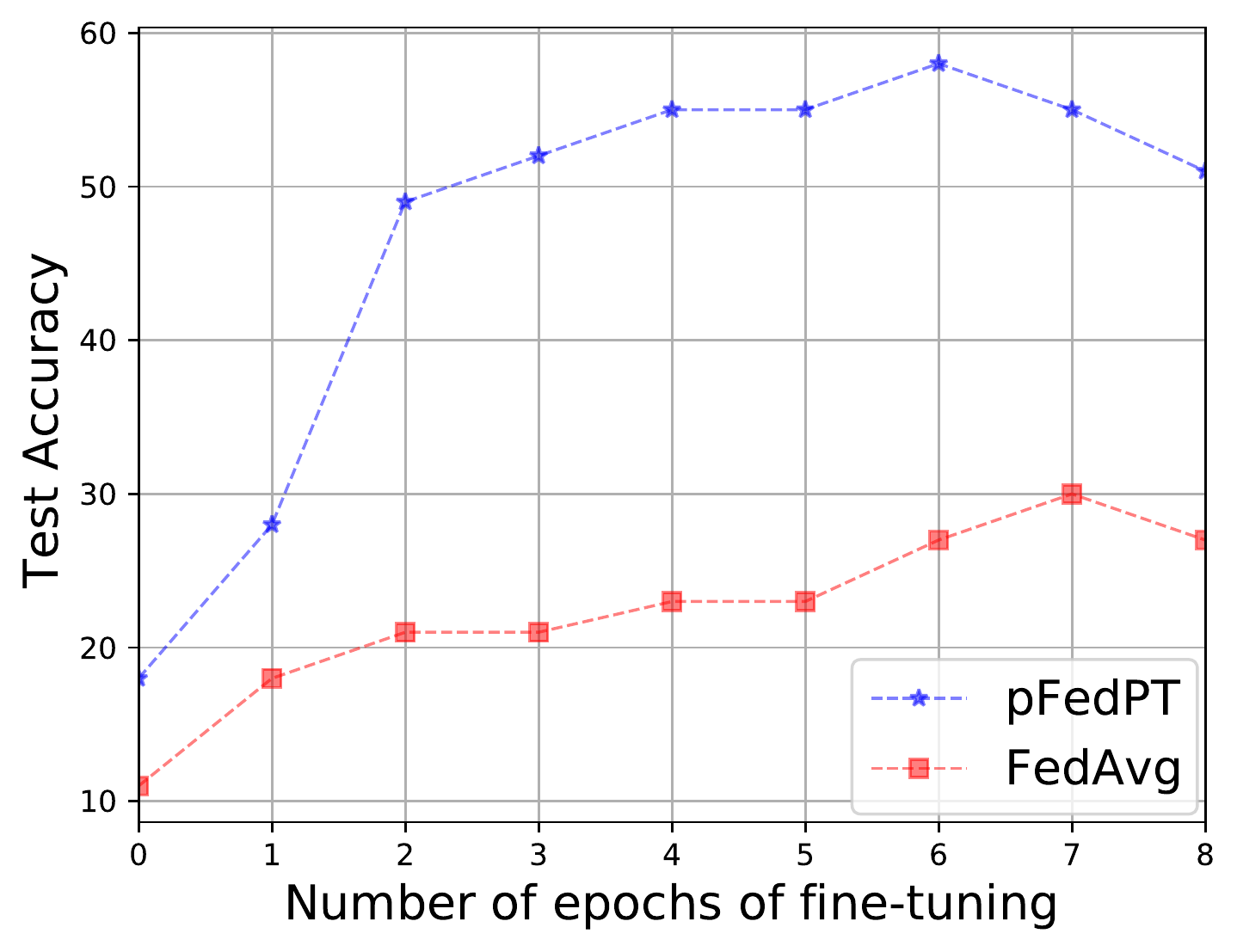}
    \captionsetup{font=small}
    \vspace{-0.8em}
    \caption{Test accuracy after fine-tuning the head of models trained on a client of CIFAR10 datasets with ViT}.
    \label{fig:fintune}
    \vspace{-1.5em}
\end{figure}

\paragraph{Generalization ability of the pFedPT.}
We evaluate the strength of the backbone learned by pFedPT in terms of adaptation to new clients. To do so, we first train the pFedPT and the FedAvg in the usual setting on the partition of the CIFAR10 dataset with 10 clients and the Dir (0.1) partition. Then, we encounter clients with data from Dir (0.3) partition of the CIFAR10 dataset. We assume we have access to a dataset of 400 samples for this new client to fine-tune. For the pFedPT, we fine-tune the prompt generator parameter over multiple epochs while keeping the backbone fixed. For the FedAvg, fine-tune the last layer of the backbone while keeping the other layers. Fig.~\ref{fig:fintune} shows that the pFedPT has significantly better performance than the FedAvg.
\end{document}